\documentclass{article}
\usepackage{spconf,amsmath,graphicx, amsfonts, amssymb}

\usepackage{enumitem}
\setlist{nosep, leftmargin=14pt}

\usepackage{mwe} 

\title{Reconstructing Retinal Visual Images from 3T fMRI Data Enhanced by Unsupervised Learning}
\name{Yujian Xiong$^{1}$\sthanks{Corresponding author: yxiong42@asu.edu} \qquad{Wenhui Zhu}$^{1}$ \qquad Zhong-Lin Lu$^{2,3,4}$ \qquad Yalin Wang$^{1}$}

 \address{$^{1}$ School of Computing, Informatics, and Decision Systems Engineering,\\ Arizona State University, Tempe, AZ, USA \\
     $^{2}$Division of Arts and Sciences, NYU Shanghai, Shanghai, China.\\
     $^{3}$Center for Neural Science and Department of Psychology, New York University, New York, USA.\\
     $^{4}$NYU-ECNU Institute of Brain and Cognitive Science, NYU Shanghai, Shanghai, China.}

\begin{document}
%
\maketitle
\begin{abstract}
The reconstruction of human visual inputs from brain activity, particularly through functional Magnetic Resonance Imaging (fMRI), holds promising avenues for unraveling the mechanisms of the human visual system. Despite the significant strides made by deep learning methods in improving the quality and interpretability of visual reconstruction, there remains a substantial demand for high-quality, long-duration, subject-specific 7-Tesla fMRI experiments. The challenge arises in integrating diverse smaller 3-Tesla datasets or accommodating new subjects with brief and low-quality fMRI scans. In response to these constraints, we propose a novel framework that generates enhanced 3T fMRI data through an unsupervised Generative Adversarial Network (GAN), leveraging unpaired training across two distinct fMRI datasets in 7T and 3T, respectively. This approach aims to overcome the limitations of the scarcity of high-quality 7-Tesla data and the challenges associated with brief and low-quality scans in 3-Tesla experiments. In this paper, we demonstrate the reconstruction capabilities of the enhanced 3T fMRI data, highlighting its proficiency in generating superior input visual images compared to data-intensive methods trained and tested on a single subject.
\end{abstract}
\section{Introduction}
\label{sec:intro}


To elucidate the intricate mechanisms of visual encoding and decoding processes in the human brain, researchers have employed various methodologies, ranging from measurements of brain activity to the application of deep learning techniques. Noteworthy contributions to the field include studies focused on the reconstruction of visual inputs from brain activity \cite{fang2020reconstructing, ren2021reconstructing, le2021brain2pix, seeliger2018generative, parthasarathy2017neural, gaziv2022self}. These investigations have significantly advanced our comprehension of the intricate relationship between neural activity and the reconstruction of visual stimuli.


Building upon recent advancements in visual representations, contemporary investigations have delved into the domain of functional Magnetic Resonance Imaging (fMRI). These studies aim to capture neural responses during carefully designed experiments where subjects engage with specific visual stimuli, such as natural scene images. The exploration spans a spectrum of datasets, from smaller-scale repositories like BOLD5000 \cite{chang2019bold5000}, Deep Image Reconstruction \cite{shen2019deep}, and the Generic Object Decoding Dataset (GOD) \cite{horikawa2017generic}, to more expansive datasets like the Natural Scenes Dataset (NSD) \cite{allen2022massive} and the recently introduced Natural Object Dataset (NOD) \cite{gong2023large}. This diversified approach enables investigations of unprecedented scale, resolution, and signal-to-noise ratio (SNR), allowing researchers to navigate the intricate landscape of reconstructing visual input images from fMRI signals.


Commencing with relatively limited sample sizes and lightweight algorithms \cite{kay2008identifying, naselaris2009bayesian}, this line of investigation has transitioned toward the intricate domain of deep learning models. Researchers are now dedicated to decoding and reconstructing visual experiences through the utilization of advanced techniques, including generative adversarial networks (GAN) and self-supervised learning directly from fMRI data \cite{ren2021reconstructing, le2021brain2pix, gaziv2022self, chen2023seeing}.  Recent studies have incorporated additional information by integrating semantic annotations of input images to augment the decoding process \cite{fang2020reconstructing, lin2022mind, takagi2023high}. This progressive shift reflects an increased commitment to capturing the intricate nuances of neural representations, emphasizing the integration of cutting-edge methodologies to enhance the reconstructed visual experiences based on fMRI.

However, most high-performance studies necessitate subject-specific training or fine-tuning from scratch and high-resolution (7T) fMRI scans to mitigate the challenges posed by individual brain variability and the considerable number of parameters in generative models \cite{lin2022mind, chen2023seeing, takagi2023high}. From the standpoint of a practical fMRI decoding framework, these approaches encounter limitations when applied to new subjects who have not undergone extensive high-Tesla experiments. This constraint restricts their practical applicability in scenarios where the model is trained on subjects with ample 7T fMRI data but is subsequently employed to decode subjects with shorter and lower resolution (3T) fMRI scans.

Addressing this challenge, we propose a novel framework designed for the reconstruction of visual stimulus images utilizing enhanced 3T fMRI data across subjects generated by the Optimal Transportation Guided GAN (OT-GAN) \cite{wang2022optimal, zhu2023optimal}. Unlike conventional approaches that rely on training and testing exclusively on single subjects with high-Tesla fMRI, our architecture is trained on a large group of subjects spanning two distinct datasets: NOD with 3T fMRI and NSD with 7T fMRI. The integrated framework demonstrates the capacity to generate enhanced 3T fMRI data of comparable quality to 7T data from subjects who have only undergone short and low-resolution 3T fMRI experiments. Notably, our approach achieves superior performance in terms of Fréchet Inception Distance (FID) \cite{heusel2017gans} and human judgment when compared to models that were trained and tested on single subject.

\section{CONTRIBUTIONS STATEMENT}
\label{sec:problem}

Our study makes the following key contributions:
\begin{itemize}
    \item \textbf{OT-GAN for retinotopic fMRI enhancement}: We introduce our OT-GAN tailored for generating enhanced 3T fMRI data from the NOD dataset in the shared space of 7T fMRI, aligning with each shared visual input.
    \item \textbf{Linear Regression Training}: We train two linear regressions on the original 7T fMRI NSD data and generated enhanced 3T fMRI data. These regressions are designed to map the collection of visual inputs and semantic annotations to their respective latent presentations, facilitating subsequent reconstruction.
    \item \textbf{Reconstruction on untrained subjects}: We leverage Stable Diffusion \cite{rombach2022high} to generate reconstructed images from 3T fMRI of an untrained subject. Our results demonstrate superior image quality and FID score compared to subject-specific methods.
\end{itemize}

These contributions collectively advance the field by proposing an effective framework for generating enhanced 3T fMRI data, training linear regressions for latent representation mapping, and employing Stable Diffusion for high-quality image reconstruction.

\section{Methods}
\label{sec:methods}
\begin{figure*}[t]
\begin{minipage}[b]{1.0\linewidth}
  \centering
  \centerline{\includegraphics[width=14cm]{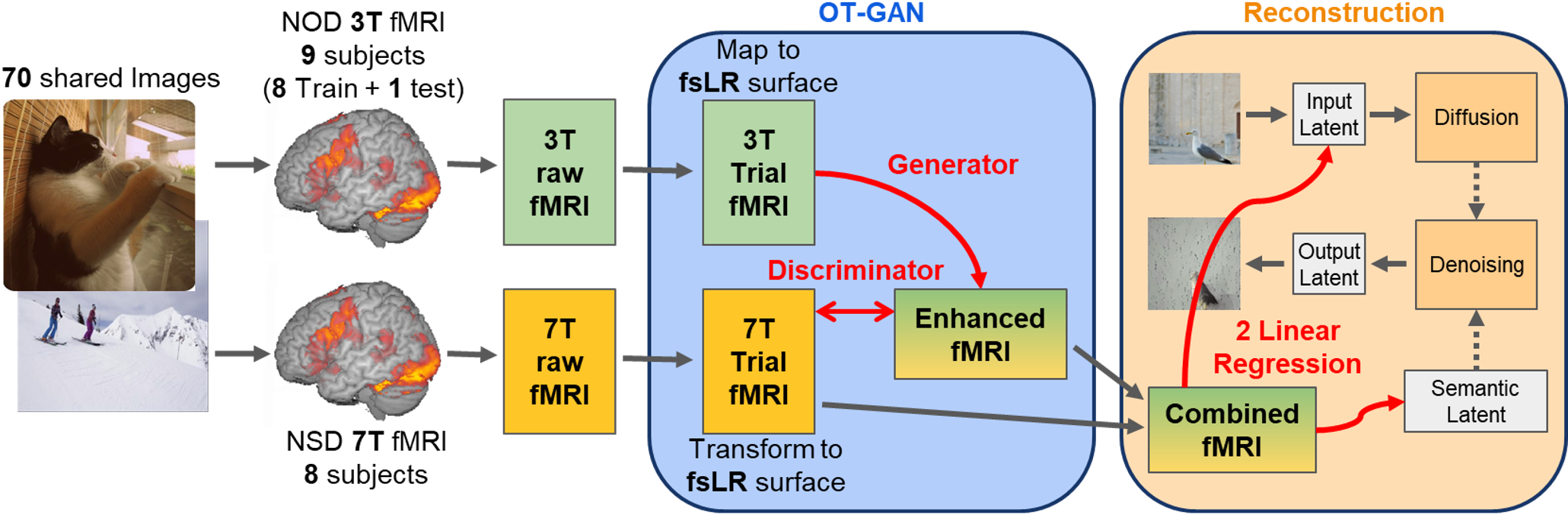}}
  \centerline{(a)}\medskip
\end{minipage}

\begin{minipage}[b]{1.0\linewidth}
  \centering
  \centerline{\includegraphics[width=14cm]{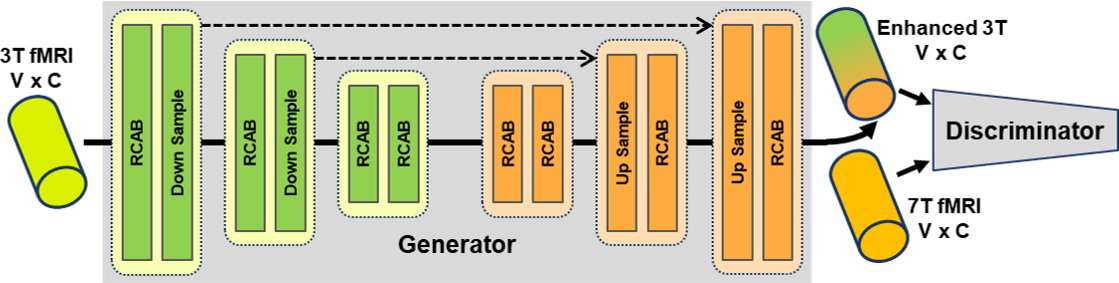}}
  \centerline{(b)}\medskip
  \vspace{-0.5cm}
\end{minipage}
\caption{Illustration of proposed framework: (a) overview of the entire pipeline, red arrows indicate models requiring training, (b) overview of the proposed OT-GAN structure.}
\label{fig:framework}
\end{figure*}

Fig.\ref{fig:framework}  illustrates the framework of our proposed method.

\subsection{Dataset and Splitting}
We employed two distinct datasets in our study: the Natural Scenes Dataset (NSD) \cite{allen2022massive} and Natural Object Dataset (NOD) \cite{gong2023large}.

\textbf{Natural Scenes Dataset (NSD)}: NSD offers high resolution 7-Tesla fMRI data from 30 to 40 sessions. During these sessions, subjects viewed natural scene images sourced from MS-COCO dataset \cite{lin2014microsoft}, which provides images with rich contextual information and detailed annotations, enhancing the depth of the visual stimuli compared to previous fMRI studies. 

\textbf{Natural Object Dataset (NOD)}: NOD consists of 3-Tesla fMRI data from 1 to 6 sessions, involving  30 subjects viewing images from both the MS-COCO and ImageNet datasets.

Our fMRI preprocessing is grounded in the assumption that subjects performing the same visual tasks yield similar fMRI recordings. We identified 70 images that were shared between the NSD and NOD experiments. This shared set enables us to extract corresponding fMRI data for training our GAN model, facilitating the generation of enhanced 3-Tesla data from the low-quality NOD records to match the 7-Tesla NSD records. For the NSD dataset, all 8 subjects contribute to our 7T data. For the NOD dataset, we selected 9 out of 30 subjects who participated in the short MS-COCO imaging session to provide our 3T data.

During the GAN training, trials related to the 70 shared images from the 8 NSD subjects are used as 7T fMRI, and trials related to the 70 shared images of the first 8 NOD subjects are used as 3T fMRI. For training the linear regression model for image reconstruction,  a combination of all 7T fMRI and enhanced 3T fMRI generated by each trial from 3T fMRI is utilized. Instead of testing on untrained images, as done in some previous studies~\cite{lin2022mind,takagi2023high,chen2023seeing},  our framework is evaluated on a new, untrained subject with low-quality fMRI (i.e., the 9th subject in NOD dataset). His 3T fMRI data is fed to the GAN for enhancement, and the enhanced fMRI data is then used to reconstruct visual input images. For a comprehensive overview, refer to the detailed dataset splitting provided in Table \ref{t1}.

\begin{table}[h]
    \centering
    \begin{tabular}{|c|c|c|}
    \hline
         Dataset & subjects & fMRI Trials* \\
         \hline
         NOD (3T) &  $9(8+1)$  &  $9 \times 10$  \\
         \hline
         NSD (7T) & $8$ & $8 \times 3$  \\
         \hline
         Train (7T + enhanced 3T) &  $8+8$  & $8\times10 + 8\times3$ \\ 
         \hline
         Test (enhanced 3T) & 1 & $10$ \\
         \hline
    \end{tabular}
    \caption{Collection of NSD, NOD dataset and train-test split, * are counting number of trials for each shared image.}
    \label{t1}
\end{table}

\subsection{fMRI Preparation}

To ensure the proper functioning of our GAN model, it is imperative that the space of 3T and 7T fMRI trials is correctly aligned. Consequently, we opted for the 32k fsLR surface space. For subjects in the NOD dataset, individual fMRI data underwent preprocessing and registration onto the 32k fsLR surface using the ciftify toolbox \cite{dickie2019ciftify}. This process involved applying a high-pass temporal filter (cutoff = $128$ s) and a spatial smoother (FWHM = $4$ mm). Further details on the preprocessing steps can be found in the dataset description \cite{gong2023large}. As for subjects in the NSD dataset, the original researchers provided preprocessed individual fMRI data on the fsaverage surface, which was subsequently transformed to 32k fsLR using the neuromaps toolbox \cite{markello2022neuromaps}. All mappings of fMRI trials are stored separately for the left and right hemispheres $V$, treated as two distinct channels $C$.

\subsection{fMRI Generation with Unpaired OT-GAN}

Denote $a \sim \mathbb P_{X}$ and $b \sim \mathbb P_{Y}$ as two probability measurements on source and target distributions. Given a loss function $\mathcal L : (x,y) \rightarrow \mathbb R_+$, we have the Monge's optimal transport mapping $G_\theta : Y \rightarrow X$ with optimal transportation cost between $b \rightarrow a$ will be:
\begin{align}
    G^*_\theta =& \inf_{f:a(x)=b(G^{-1}_\theta(x))} \int_Y \mathcal L (y, G_\theta(y)) db(y) 
\end{align}
\begin{align}
    \theta^* =& \mathop{\arg\min}\limits_{\theta:\mathbb P_X = \mathbb P_{G_\theta (Y)}} \mathbb E_{y \sim \mathbb P_Y} \mathcal L (y, G_\theta(y)) \label{e2}
\end{align}

The parametric optimal transport map could be considered as a deep neural network with a well-defined unsupervised training schedule on unpaired data. Equation \ref{e2} can be relaxed to an unconstrained optimization by adding a Lagrange multiplier term $\lambda d(\mathbb P_X, \mathbb P_{G_\theta(Y)})$, where $d$ corresponds to the divergence between two distributions with all element smaller than $0$. We can calculate the adversarial divergence by Wasserstein-1 distance \cite{wang2022optimal} using $D_\mu$ as a parametric discriminator: 
\begin{align}
    \mathbb W_1 (\mathbb P_X, \mathbb P_{G_\theta(Y)}) = \sup_{||D_\mu||\leq 1} &\mathbb E_{x \sim \mathbb P_X} (D_\mu(x)) - \nonumber \\
    &\mathbb E_{y \sim \mathbb P_Y} (D_\mu(G_\theta(y)))
\end{align}

In summary, high alignment between low-quality and high-quality fMRI could be achieved by trading off consistency between enhanced and high-quality fMRI. The two parametric mappings $G_\theta$ and $D_\mu$ could be trained through a GAN structure by optimizing the combined objective function:
\begin{align}
    \max_\theta \min_\mu E_{y \sim \mathbb P_Y} \mathcal L (y, G_\theta(y))  + \lambda \mathbb W_1 (\mathbb P_X, \mathbb P_{G_\theta(Y)})
\end{align}

We choose the conventional MSE as our loss function $\mathcal L(y, G_\theta(y)) = \mathbb E ||y - G_\theta(y) ||^2$. To maintain the hidden structural information in the original fMRI, we use a U-shape Network to pass underlying information across down-sample layers. Residual Channel Attention Blocks (RCAB) \cite{wang2022optimal, zhu2023optimal} are modified to 1D convolution and then used to replace naive connections. Down-sample and up-sample blocks are scaling the input dimension by $\times1/2$ or $\times2$, respectively. And the discriminator is adapted from \cite{ledig2017photo}.

\subsection{Image Reconstruction based on Diffusion Model}

The Diffusion Model (DM) \cite{ho2020denoising} is a probabilistic generative model that recovers sampled variables from Gaussian noise to a specific learned distribution through iterative denoising. In each diffusion step, the input $\boldsymbol{x}$ is blurred by the addition of small Gaussian noise $\epsilon$ iteratively over time points $t$. This process can be summarized as $\boldsymbol{x}_t = \sqrt{\alpha}x_0 + \sqrt{1-\alpha_t}\epsilon_t$. The denoising step employs a neural network $f\theta (\boldsymbol{x}_t,t)$, typically a U-Net, to restore the previous input. The learning objective converges to $f\theta (\boldsymbol{x}_t,t) \simeq \epsilon_t$ \cite{ho2020denoising}. In the Latent Diffusion Model (LDM), an encoder-decoder pair is applied before and after the diffusion-denoising network, accelerating the process by transferring the original pixel space calculation into a smaller encoded latent space.

Recent studies \cite{rombach2022high} have demonstrated that this method can be improved by incorporating conditional input into the neural network, such as semantic annotations of the original input, to generate high-quality natural images. Given the performance of Stable Diffusion \cite{rombach2022high}, it becomes an ideal choice for testing our enhanced 3T fMRI data in visual image reconstruction tasks.

Drawing inspiration from a recent study's straightforward training framework \cite{takagi2023high}, we utilized the pre-trained Stable Diffusion v1.4 and kept all its weights frozen. Two linear neural networks were applied to our combined fMRI training set to generate the latent representation of corresponding visual images and their annotations, respectively. After training, the enhanced 3T fMRI from the 9th subject of the NOD dataset was used for generating test images. The enhanced fMRI from his ten 3T scan trials when viewing each image was averaged and then applied to the two networks. The generated latent representation of visual and semantic inputs was subsequently used for reconstruction in the pre-trained Stable Diffusion.

\vspace{-0.1cm}

\section{Results}
\label{sec:results}

\begin{figure}[htb]
\begin{minipage}[b]{0.99\linewidth}
  \centering
  \centerline{(a) Ground Truth}
  \centerline{\includegraphics[width=2cm]{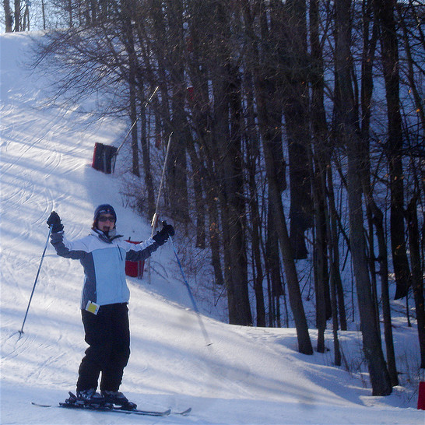}
              \includegraphics[width=2cm]{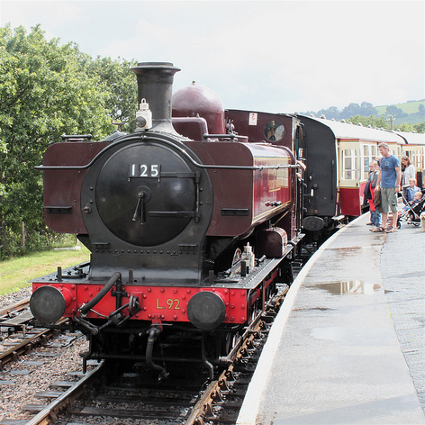}
              \includegraphics[width=2cm]{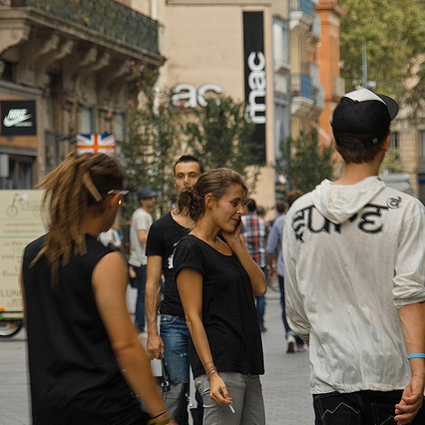}
              \includegraphics[width=2cm]{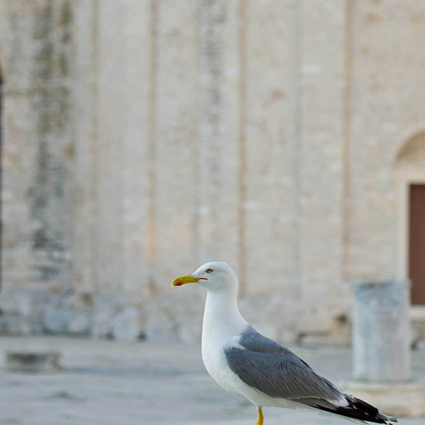}}
  \centerline{}
\end{minipage}
\begin{minipage}[b]{0.99\linewidth}
  \centering
  \centerline{\includegraphics[width=2cm]{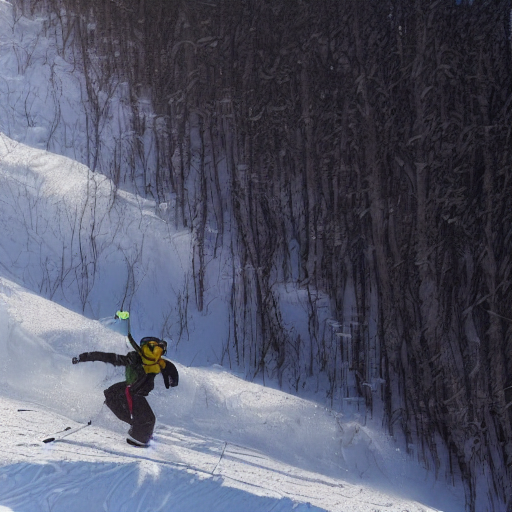}
              \includegraphics[width=2cm]{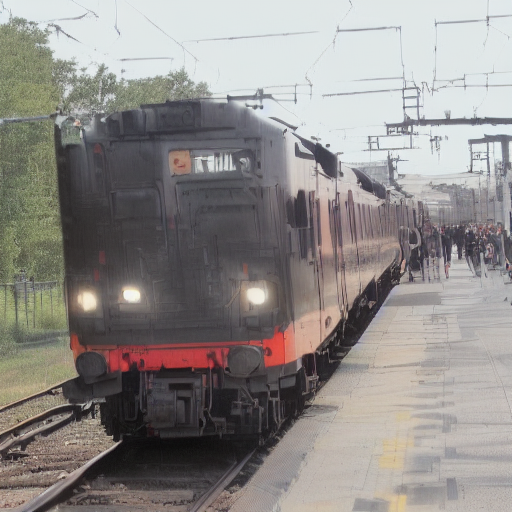}
              \includegraphics[width=2cm]{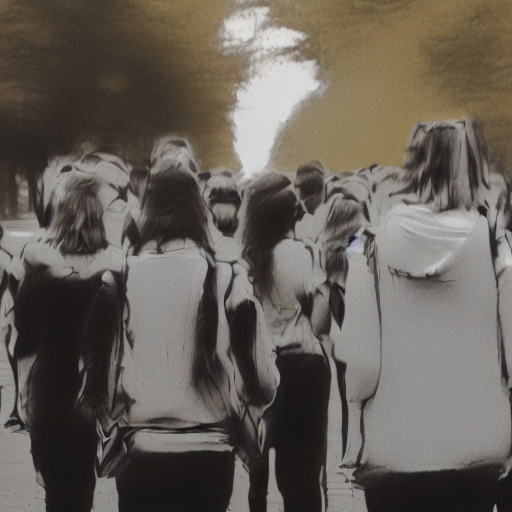}
              \includegraphics[width=2cm]{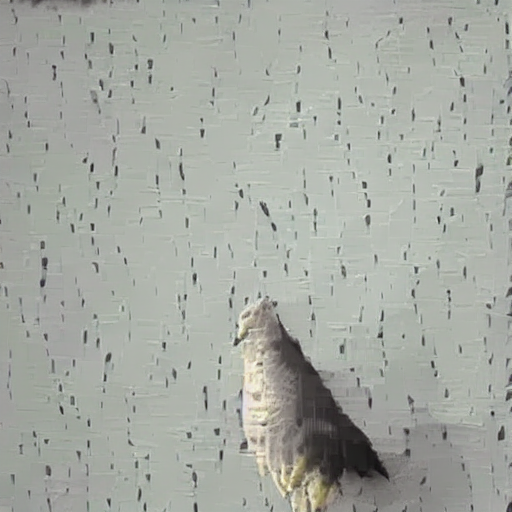}}
  \centerline{(b) Reconstructed Image}
  \vspace{-0.4cm}
\end{minipage}
\caption{Ground truth and corresponding reconstruction from the 9th subject of the NOD dataset.}
\label{fig:result}
\end{figure}

Our reconstruction are displayed in Figure \ref{fig:result}. For each input visual images, we generated five different reconstruction, and we present the one with the least FID. The FID \cite{heusel2017gans} is a widely used metric for assessing image generation quality by measuring the distance between ground-truth images and reconstructions. Compared to recent reconstruction methods \cite{lin2022mind, chen2023seeing, takagi2023high}, our framework produces considerably high-quality natural scenes with a test subject that hasn't been included in training.

We assess our performance using FID across the entire set of 70 shared images. The metric serves to compare our results with two studies that focus on NSD dataset: Lin \textit{et al}. \cite{lin2022mind} and Takagi \textit{et al}. \cite{takagi2023high}, as outlined in Table \ref{t2}. Our FID scores demonstrate the superiority of our approach over previous methods, which were trained and tested on a single subject with a lengthy 7T fMRI scan. Notably, our framework excels when tested on an untrained new subject who has only undergone a light-weight 3T fMRI experiment.

\begin{table}[h]
    \centering
    \begin{tabular}{|c|c|}
    \hline
         Method &  FID \\
         \hline
         Lin \textit{et al}. &  $45.02$   \\
         \hline
         Takagi \textit{et al}. & $52.14$  \\
         \hline
         \textbf{Proposed} &  $\textbf{40.48}$   \\
         \hline
    \end{tabular}
    \caption{Comparison of FID score between NSD reconstruction on 70 shared images}
    \vspace{-0.5cm}
    \label{t2}
\end{table}

\section{Conclusion}
\label{sec:conclusion}

In this study, we introduced a robust OT-guided unpaired GAN model designed to generate enhanced 3T fMRI scans. These scans exhibit comparable quality and reconstruction performance to native 7T fMRI scans, as demonstrated across two distinct datasets. The enhanced 3T fMRI data effectively preserves both visual and semantic information, subsequently employed in a straightforward linear regression model for visual imaging reconstruction utilizing Stable Diffusion.

Our approach underscores its capability to undertake reconstruction tasks requiring extensive and high-quality fMRI scans, particularly for new subjects with only brief and lower-quality fMRI scans. Through the adept utilization of a well-trained GAN model, our method proves promising in addressing such challenges. In subsequent work, we envisage modifying our framework to address complex tasks, such as solving population receptive field (pRF) mapping and facilitating brain disease diagnoses, all within the application of brief 3-Tesla fMRI experimental settings.


\section{COMPLIANCE WITH ETHICAL STANDARDS}
\label{sec:ethical}

This research study was conducted retrospectively using human subject data available in open access by \cite{lin2014microsoft, allen2022massive, gong2023large}. Ethical approval was not required as confirmed by the license attached with the open-access data.

\section{Acknowledgments}
\label{sec:acknowledgments}
The work was partially supported by NSF (DMS-1413417 \& DMS-1412722) and NIH (R01EY032125 \& R01DE030286).


\bibliographystyle{IEEEbib1}
\bibliography{OTfMRI}

\end{document}